# Decision-Analytic Approaches to Operational Decision Making: Application and Observation

Tom Chávez[1]

1. Department of Engineering-Economic Systems, Stanford University
Sun Microsystems, Inc, Menlo Park CA email: tommyc@corp.sun.com

*Abstract.* **Decision analysis (DA)** and the rich set of tools developed by researchers in **decision making under uncertainty** show great potential to penetrate the technological content of the products and services delivered by firms in a variety of industries as well as the business processes used to deliver those products and services to market. In this paper I describe work in progress at Sun Microsystems in the application of decision-analytic methods to Operational Decision Making (ODM) in its World-Wide Operations (WWOPS) Business Management group. Working with members of product engineering, marketing, and sales, operations planners from WWOPS have begun to use a decision-analytic framework called **SCRAM (Supply Communication/ Risk Assessment and Management)** to structure and solve problems in product planning, tracking, and transition. Concepts such as **information value** provide a powerful method of managing huge information sets and thereby enable managers to focus attention on factors that matter most for their business. Finally, our process-oriented introduction of decision-analytic methods to Sun managers has led to a focused effort to develop decision support software based on methods from decision-making under uncertainty.

## 1.0 Operational Decision Making at Sun Microsystems

Sun Microsystems, Inc., is a large, established provider of technology products such as workstations, servers, printers, and computer storage devices, with annual revenues of nearly $7 billion dollars. Balancing supply and demand for Sun's products is difficult given the overall volatility of the technology marketplace, complex dynamics and uncertainties in customer preference, and extremely short product lifetimes—less than 18 months for most technology products. In such a fast-paced business environment, an efficient, intelligent execution capability is critical to long-term business health. The wing of the company charged with building and maintaining overall execution capability is called World-Wide Operations (WWOPS), an organization comprising several large groups such as Supplier Management, Manufacturing, Logistics, Product Engineering, and Business Management.

WWOPS Business Management is responsible for longer-term business planning related to the design, manufacturability, and delivery to market of Sun's products. It is not a tactical or logistics organization. Rather, it works with other groups such as Engineering, Marketing, and Sales to gather and manage assets to support introduction of new products, end-of-life transitions for older products, and supply/demand balancing for what are called sustaining products. The translation of executive-level decisions — to pursue particular product lines or to size existing businesses in particular ways, for example — into achievable plans which exhaustively take account of supply, technological, and market constraints poses a formidable challenge in operational decision making.

Consider a quick case study: many believe that Apple's inability to measure and meet demand for its PowerPC line of computers during the fiscal 1995 year was the first and possibly largest blunder leading to its long and dangerous downturn. Because Apple thoroughly misjudged demand for PowerPC's, and because it had not positioned supply of components far enough in advance to adjust to the demand upturn, it suffered a prolonged beating in the marketplace as customers turned to competitors' products. Consequently, Apple's market share fell from a 1992 peak of 11.2% to 6.7% as of the end of calendar year 1995, according to Dataquest, Inc. As Apple's difficulties vividly demonstrate, the stakes on technology product supply decisions can be extremely high.

In this paper I describe work in progress at Sun in the application of decision-analytic (DA) methods to operational decision making (ODM) in its WWOPS Business Management group. Working with members of product engineering, marketing, and sales organizations, operations planners have begun to use a decision-analytic framework called **SCRAM (Supply Communication/ Risk Assessment and Management)** to structure and solve problems in product planning, tracking, and transition. Concepts such as **information value** provide a powerful mechanism for probing and managing huge information sets and thereby enable managers to focus attention on factors that matter most for their business. Finally, our process-oriented introduction of decision-analytic methods to Sun managers has led to a focused effort to embed these decision making methodologies inside Business Management's information technology (IT) architecture. I'll briefly describe an effort underway to build decision support software based on methods from decision making under uncertainty to help manage Sun's large portfolio of technology products.

## 2.0 SCRAM

In this section I describe in an anecdotal way the evolution of SCRAM and the lessons I've learned in bringing it online as a process-oriented framework for ODM at Sun.

### 2.1 Rationale for SCRAM

In the past, different groups within Sun's WWOPS organization funded efforts based on Operations Research (OR) techniques to address problems in operational strategy and planning. Managers had become disenchanted with such approaches, however, because they felt that those approaches were providing sophisticated answers to the wrong problems. They generally felt that OR tools abstracted away many important characteristics of the problems those tools purported to solve. Moreover, because managers did not understand the inputs to those OR models, they distrusted the outputs.

In some cases the consultants who built the models suggested that understanding how the models worked was really beyond managers' expertise. Managers achieve their positions in part because they are extremely self-confident, and so it seems ill-advised to suggest that it is beyond their ability to understand a project they themselves have funded. Such an explanation seems especially wide of the mark, moreover, because Sun is in essence a technically-minded company. Many managers rise through the ranks from engineering groups and are more than adequately equipped to understand the gist of a mathematical model.

The Vice-President of Business Management, an engineer by training, thought about many of the diverse activities happening inside his organization—e.g., strategies for new product introductions, end-of-life transition plans, processes for asset management— and he came to a simple realization. He saw that a core theme ran through all of the activities in his group, and that that theme was in many cases inadequately supported by many of the OR techniques WWOPS had tried using in the past. As he succinctly put it, "*Decision making* is the core of what we do. So don't just give me another model. Give me enhanced decision making capability."

### 2.2 Development of SCRAM

I was challenged to assess the state of the art of supply planning and management at Sun and to propose new processes, methods, and tools for doing it better. I began by interviewing a group of Platform Managers and Key Component Managers to understand the key challenges before them. Platform Managers are responsible for developing supply strategies for desktop and server products. Key Component Managers manage critical components which are needed to make Sun's products or sold separately as peripherals (e.g., memory chips, storage devices). Three clear themes quickly emerged.

1. **Communication**: Planners must communicate with members of diverse groups such as Product Engineering, Marketing, and Sales to understand a very broad array of issues shaping the design and delivery of Sun's products in addition to the dynamics of market demand for those products.

2. **Credibility**: Using data available in large databases and the information gathered through collaboration with members of other groups, planners must design credible long-term supply strategies which balance Sun's worldwide supply/demand equation.

3. **Complexity**: Planners have to build plans which explicitly take account of a wide range of complex and mostly uncertain factors under strict time constraints. In many situations, planners' initial reaction to such complexity is to say that an issue is "totally chaotic" or "completely random—there's really no way to get your head around this."

I began to tell planners and analysts about decision analysis and utility modeling. I assumed that the connections between what I was telling them and the problems they faced would be immediately apparent. I was immediately and not-so-gently reality-checked. As one of the planners told me, "You're offering me caviar and champagne when I really need bread and water." Lesson #1 immediately sunk in:

**Lesson #1**: *UA/DA methods, if they are to be usefully applied in an organization, need to be wrapped inside the issues and concerns immediately relevant to the organization. The problem context comes first; methodology follows.*

I quickly realized that I would have to change my approach. I began browsing internal home pages to learn about Sun's products. I badgered platform and key component managers for minutes to meetings, product specifications—anything that would help me to learn the nitty-gritty details of Sun's products. I also began collecting what I had heard in my interviews into a framework which could describe the state of the art for supply planning while also suggesting ways of doing it better. The result—or, more accurately, the work in progress—is SCRAM. SCRAM is a collection of three things: a **glossary** of shared terms; a **set of simple distinctions** for categorizing the elements of supply plans; and finally, a library of agreed-upon **knowledge maps and influence diagrams** for describing the structure of recurring challenges and problems.

In fact, SCRAM is nothing more than a simple wrapper for decision analytic methods applied to ODM at Sun. By taking inventory of the essential factors related to supply planning—e.g., market growth, supply quality, customer migration—SCRAM provides a relatively stable vocabulary to enable planners to communicate more effectively and consistently with experts in other parts of Sun's organization. The purpose of introducing distinctions such as Supply vs. Demand and Inside Sun vs. Outside Sun is to enable planners to understand their environment better and to establish clearer patterns for thinking about risk and uncertainty. Belief nets and influence diagrams enable planners to formulate their problems using simple graphical models with qualitative flavor. Taken together, these three elements provide a basis for the development and rigorous analysis of short- and long-term supply strategies.

### 2.3 Dealing with uncertainty

While I had some initial success in gathering those elements into a simple framework with a cute name, I still faced significant obstacles. After some of my initial presentations to managers about SCRAM and its potential uses, my director told me that "SCRAM seems like nice conceptual framework, but we need to see how you actually use this stuff." For example, I had talked about the idea of using a simple graphical model to describe uncertainty and I had given examples of belief networks and influence diagrams, but people had not really internalized the relevance or power of completing such an exercise.

Working with a receptive Platform Strategy Manager, I decided to attack one of the central problems Sun was facing at the time, the introduction of our new UltraSPARC line of workstation computers. UltraSPARC was a new computing paradigm, driven by the design of a new Ultra microprocessor chip with significantly higher computing speed; a new system model called UltraPort Architecture (UPA), which specified a new protocol for communication between the CPU and external graphics, network, memory, and storage devices; and an advanced graphics capability called Creator 3D Graphics. The transition to UltraSPARC was a focused, company-wide commitment. The dynamics and complexities of the move carried huge risks.

The central difficulties in any product transition include

- assessing the pace and pulse of market demand using scarce market information—scarce because we do not yet have experience selling the new product;
- monitoring the rate of migration in the customer base from sustaining products to new products;
- positioning assets for sustaining and new products to exploit commonalities in shared components and to manage uncertainties and constraints related to the use of new or unique components;
- integrating knowledge from diverse corners of the organization to make tough operational decisions while simultaneously supporting company objectives—this involves distinguishing between what we would like to happen (our goal) from our best assessment of what is most likely to happen (our forecast); and
- making trade-offs to (a) minimize risk in excess inventory of highly expensive components such as microprocessors, (b) maximize overall revenue, (c) preserve the stability of our supplier network, and (d) keep customers satisfied.

I completed an intense knowledge engineering cycle with the Platform Manager in an attempt to make sense of some of the divergent signals we were receiving at the time from Marketing and Sales groups. It was interesting to observe the effects of **motivational and availability biases** [Tversky and Kahneman, 1974] when we approached those sources for more refined information. For example, it became clear that current products such as Sparc 5's and Sparc 20's were familiar not just to our customers but to the salespeople selling to our installed base of customers. Ultra is based on a new design and therefore required a new and mostly untested sales pitch. Additionally, salespeople receive bonuses based on the difference between baseline targets and actual sales, so they have a strong incentive to set the bar low. Marketing people, meanwhile, are congenital optimists. High-tech marketers do not advance in their careers by being skeptical of the technologies they promote; their professional success depends heavily on their ability to convince customers and other members of the corporation that the products they help manage are "best-of-breed," in high-tech parlance.

I realized that Marketing and Sales groups were operating not simply from different assumptions but from radically different perspectives. Business Management's objective in this situation was to establish rules for clear communication so that we could understand and diagnose the source of disagreement. Belief networks provided an excellent methodology for depicting the elements of each perspective. They encouraged people to talk about the problem using a common representation with simple semantics. When skeptics argued that I was simply drawing a "mind map" or a "flow diagram," I showed them how the knowledge map could be used analytically to capture a rich structure of probabilistic and functional dependencies. A few engineers working in product management and marketing found this approach particularly interesting and were inclined to give the results of our analyses a high level of credibility because the solution methodology appealed to their sense of technical rigor.

It is important to emphasize Business Management's role in such an environment: while we are ultimately accountable for the Ultra supply strategy, several different groups were also responsible. Lesson #2 became apparent.

**Lesson #2:** *In ODM, the notion of a single Bayesian decision maker is an abstraction of limited use. A more realistic approach takes account of what I will call the **decision participants**. Capturing their information and knowledge and explicitly including them throughout the decision-making cycle appears critical for the successful application of decision-theoretic methodologies in an organizational context.*

Figures 1 shows a belief network for the Ultra forecast from the Sales perspective. Bold-bordered nodes represent deterministic nodes.

**FIGURE 1.** Salespeoples' belief net for Ultra forecast.

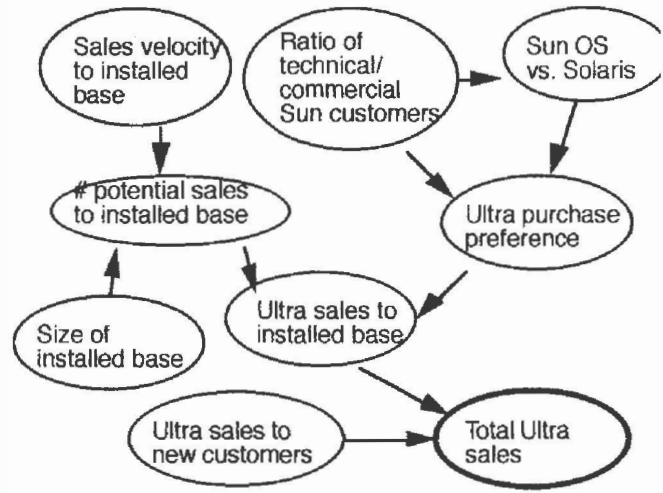

An important advantage of formulating the belief network model shown above was to identify and make explicit the dynamics of Ultra sales in terms of different customer profiles. There was a general feeling that Ultra sales would be constrained by the starting configurations of the computing environments in which users were already configured. Since UltraSPARC requires installation of the Solaris 2.5 operating system, different salespeople felt that this would be a strong gating factor on our ability to generate Ultra sales. Customers are slow to upgrade their operating systems (OS) because doing so in the past required forward-porting most of their basic applications. Though one of the advantages of Solaris 2.5 was full binary compatibility with most existing applications, the perception of difficulty in moving to a new version of the OS persisted. Moreover, salespeople believed that the migration path would be harder for customers who were using the older SunOS operating system, while customers using Solaris 2.x (i.e., Solaris 2.0 - Solaris 2.4) had a higher chance of upgrading to Ultra.

After more discussion, salespeople agreed that these factors were fundamentally shaped by whether or not the customer was a technical user or a commercial client. Their intuition was that commercial customers running SunOS would be the slowest movers to Ultra. The probabilistic structuring of conditional dependencies using a belief network also allowed us to integrate hard data (e.g., size of the installed base) with more subjective assessments (e.g., Ultra purchase preference).

Finally, we used the preceding model to sanity-check forecasts of Ultra sales. In some cases, there were salespeople who had not participated in the knowledge engineering cycle but still submitted forecasts on total Ultra demand. If they submitted an Ultra forecast and quickly agreed to the assessments driving the other parts of the model, then the number of new Ultra sales to customers outside the installed base became a matter of inference, as depicted in Figure 2. If the resulting distribution on Ultra sales to new customers lacked credibility, then the salespeople with the new forecasts were challenged to defend their numbers by re-examining their assumptions about the other elements of the model, e.g., providing a reason why the purchase preference of the average Ultra customer might be different from the assessment they had previously agreed to.

**FIGURE 2.** Sanity-checking the Ultra forecast.

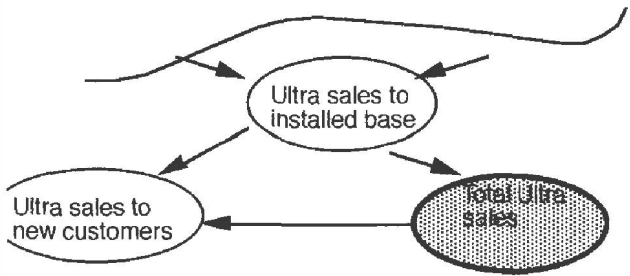

Figure 3 shows a belief network for Ultra sales from the Marketing perspective.

**FIGURE 3.** Marketing's belief net for Ultra sales.

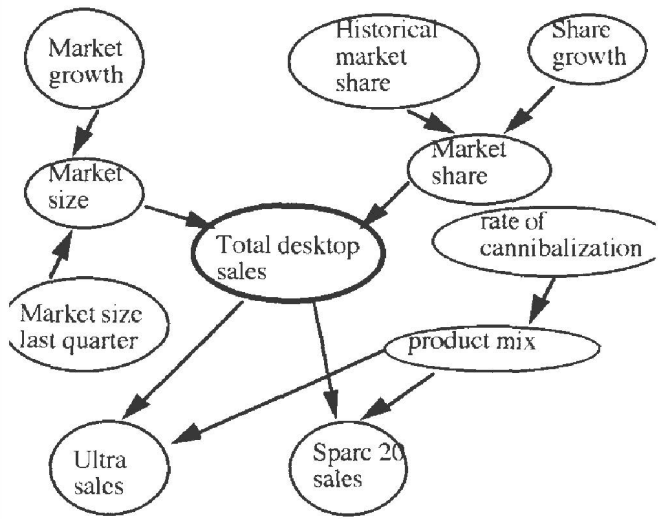

As evinced in Figure 3, marketing people examine the market from what seems to be a totally different level of abstraction. They generally analyze broader factors such as market size and market growth. In this case they also found it useful to assess Ultra sales relative to a drifting or cannibalizing effect between the current high-end product, Sparc 20, and the newly available Ultra.

Before using these tools, Marketing and Sales groups had no clear, directed means of making sense of their divergent forecasts. The knowledge mapping approach gave Business Management an opportunity to defuse—at least partially—the passionate disagreements between the two sides by eliciting their assumptions, depicting them in evocative graphical models, and performing the necessary analysis to derive more credible forecasts. Lesson #3 is an important one.

**Lesson #3:** *Uncertainty models such as belief networks can be used in ODM on a fast, flexible basis to provide structure in situations characterized by ambiguity and disagreement. Evocative models such as these provide decision participants with a high level of clarity and consensus by forcing them to say what they mean and to mean what they say.*

From this point is was possible to extend the belief network models into an influence diagram to model the business decision of how to position assets to respond to uncertain demand for Ultra. The analysis incorporated cost and margin information regarding both low- and high-end products. It suggested strategies for positioning assets over different periods of time, the structuring of good contracts with suppliers, and the identification of important variables to track as the transition continues. For proprietary reasons it is not possible to discuss these developments further.

**FIGURE 4.** Integrating the forecast for Ultra demand into an influence diagram model for Ultra supply strategy.

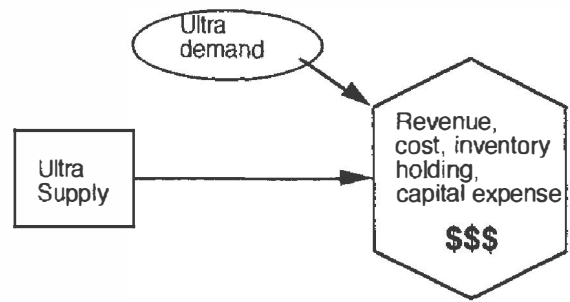

It has been especially useful to gather these models into an evolving library of models for SCRAM as a means of recording institutional rationale and of providing an audit trail on operational decisions. We have examined other product transitions recently using the SCRAM framework with positive results; I am now leading a team which is working cross-functionally to embed this approach into the basic product planning, tracking, and transition cycle of the Storage Business Unit. In contrast to the fairly static OR methods which Business Management had tested before, these decision-analytic approaches stand apart because they are dynamic and extensible. Managers and analysts who are accustomed to OR thinking worry that if they raise a new concern or introduce a new variable, then the entire house of cards will come tumbling down. It has been a challenge to get the users of SCRAM to stop searching for "the model" [Laskey, 1996] and to understand that the modeling possibil-ities with a UAI/DA approach are, practically speaking, limitless.

**Lesson #4**: *The SCRAM framework for ODM has begun to provide enhanced decision-making capability inside Sun's organization on a cross-functional basis. The chief benefits of SCRAM include improved communication among groups, increased credibility of our operational decisions and critical business forecasts, and a simple, coherent mechanism for coping with complexity. The belief networks and influence diagrams gathered inside the SCRAM library are extensible insofar as they provide modeling "chunks" which can be joined and rearranged very quickly to support rapid model-building and analysis. Such an approach is absolutely necessary to support Sun's dynamic, rapidly evolving business.*

**FIGURE 5.** The SCRAM approach creates a growing core of operational knowledge for better business decision-making.

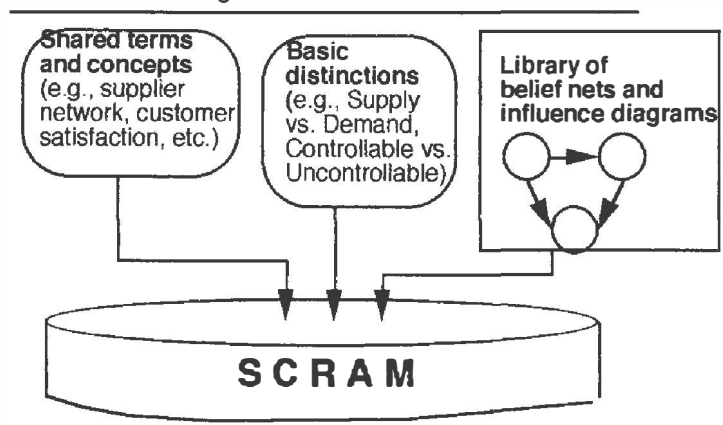

## 3.0 Information management and information value theory

One of the organizing principles Business Management has recently adopted for itself is that it is an **information brokerage**. Directors in WWOPS Business Management recently returned from their annual, two-day meeting where they assess the state of the business and plan the central initiatives and directives for the coming year. By the end of this meeting they had converged on a simple, unifying slogan: "Information is product."

While simple, the statement aptly characterizes many of Business Management's activities. Members of Business Management collect and synthesize information from different corners of the company and from various external sources, use it to make better portfolio planning decisions, and pass it on to other decision makers in the organization. Having embraced the metaphor of the information brokerage, executives began to understand the relevance and power of **value of information (VOI)** as a means of managing and valuing ever-increasing information sets.

The Ultra transition again provided a fertile ground for testing and applying this important concept. The final model that evolved after the knowledge engineering cycle with the decision participants had 12 uncertainties. Using the algorithm developed in [Chavez and Henrion, 1994] and [Chavez, 1996], I was able to analyze the value of perfect and partial information on all of those uncertainties. The concept elucidated in [Chavez, 1996] of a **Relative Information Multiple (RIM)** provides a flexible means of describing and measuring partial information value. In the Ultra problem, saturation effects in RIM curves helped us to identify the points beyond which further information-gathering was essentially overkill. It was useful to see some of the decision participants' intuitions confirmed with the VOI analysis; yet the VOI numbers also emphasized the importance of factors which were previously thought to be of lesser consequence.

We have used VOI analysis at Sun as a means of identifying areas in which it might be useful to

- hire consultants;
- search through a database;
- extend the conversation with a knowledgable expert;
- poll customers;
- refine the model;
- hire new people; or
- acquire market research.

We have performed several SCRAM analyses now, and each time I have provided VOI measures on critical uncertainties. I have observed an unexpected benefit from this kind of analysis: when provided with VOI measurements, managers feel an extra incentive to gather information, hire outside help, or talk to other people within the organization *because they had a dollars-and-cents measure of the importance of doing so*. Resolving the uncertainty and adjusting their supply strategies in response to new information gives them a more precise, personal sense of the value they add to Sun's business.

**Lesson #5**: *In an information-rich context such as ODM at Sun, the concept of VOI and supporting algorithms for its estimation in large decision models provide an effective mechanism for managing ever-expanding information sets. It also gives information-gatherers in organizations a more precise, personal sense of the importance of reducing or resolving uncertainty on variables significant to their business decisions.*

**FIGURE 6.** Information in the form of distributed expertise inside the organization, data residing in large data warehouses, plans and forecasts of other groups, or market signals from the external business environment can be used to make better decisions in principle; concepts such as VOI and efficient algorithms for estimating VOI help decision makers make more intelligent use of that information in practice.

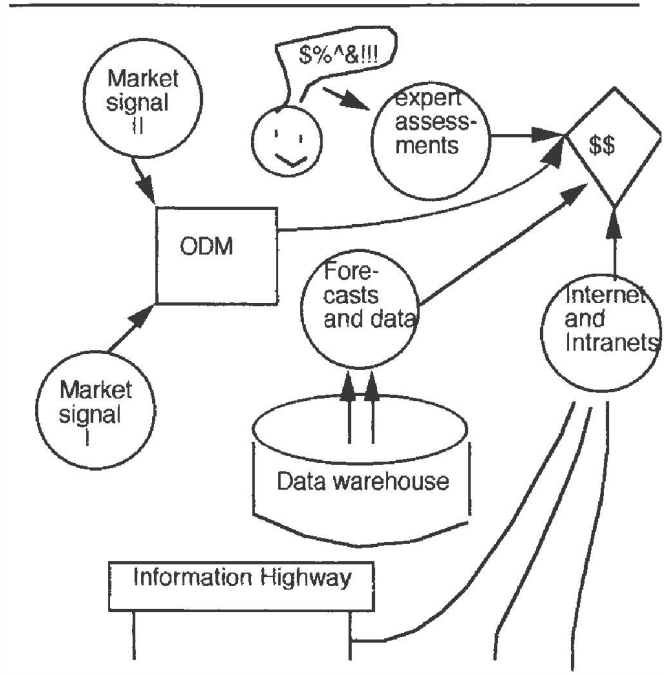

## 4.0  Development of Decision Support Software

In this section a briefly describe the thrust of a new effort to develop a decision support software environment for WWOPS Business Management at Sun.

### 4.1  Process first, tools second

I believe that management's disillusion with OR techniques had created a barrier in their willingness to try other analytic methods, and this in turn has slowed the pace of software development based on more promising approaches. My original preference would have been to begin building computer systems for decision support immediately. Yet the development of SCRAM and the application of concepts such as information value *before* investing large amounts of money in decision support has born fruit. Management has become aware of the power of these kinds of methods and then asked on their own if it would be possible to build tools to support our evolving decision-making processes.

Having first developed a clear method for formulating and solving problems via frameworks such as SCRAM, we now have a clear rationale and much greater consensus about the need for investment in accompanying tools for decision support. Echoing the Vice-President's original request, the point here is to deliver enhanced decision-making capability, not just another model. I conjecture that, if any software tool for decision support is to be really useful inside a large company, *the tool's interface must be in synch with that organization's processes*. At Sun, given initial success so far in getting people to think in terms of belief networks, the development of software tools with a belief network or influence diagram interface will from their perspective appear to be the clear, natural choice. In other words, there is a higher chance that users will use and *continue to use* the tools we are now developing because the tools look and feel like the decision-making processes they are intended to support.

**Lesson #6**: *To insure lasting success, the development of software to support business decision making must be in synch with the processes used to run the business. The interfaces for decision support software tools must match the organization's process language.*

A second benefit to the "process first, tools second" approach has been to develop an extremely precise view of the functionalities we need to embed in new software to support decision making in WWOPS Business Management. Certain methods developed in decision making under uncertainty research, while perhaps useful in other areas, will be less useful in our problem domain. For example, managers and analysts are very comfortable speaking in terms of '80% chances' and 'even odds.' I think this has much to do with the culture of the organization—Sun is, as I've already noted, a technically-minded company. For example, marketing people create roadmaps to describe the migration of customers from product $X$ to product $Y$ as percentage flows from $X$ to $Y$; this essentially requires the notion of an average customer calling a Sun sales rep and displaying some probability $p$ of requesting $X$ and a probability $(1-p)$ of requesting $Y$. In fact it has been surprising for me to see how many of the plans and reports generated within the company present results which are understandable in explicitly probabilistic terms. Thus in our environment there is less need for qualitative approaches to probability (e.g., semantic mappings of "more" or "less" likely, order-of-magnitude methods) because the business language already accommodates subjective probabilistic assertions of chance.

### 4.2  Data-mining

While our ultimate goal is to create a decision support layer for WWOPS which employs methodologies from research in decision making under uncertainty, in fact our first steps in this area must focus a bit less on decision making and more on information synthesis. Sun has millions of rows and columns of data on customers, suppliers, parts, platforms, bookings, backlog, billings, and shipments. Currently, much of this data is not really used to drive decision making; acquiring it requires knowledge of intricate data schema and the laborious development of database queries. The most important task before us is to create a more seamless, powerful way of getting at the data and mining it for valuable information about product trends and market pulse.

I have used SCRAM as a mechanism for introducing analysts at Sun to new, more powerful models for forecasting product demand and component consumption. Such models may rely on direct assessment; where possible, however, we would also like to put those millions of rows and columns of data to work. Recently developed methods for **blending statistical forecasting with belief net inference** [Dagum and Galper et al, 1995] provide an attractive means of turning our data into information to drive business decision-making. We have completed a small pilot project drawing on such techniques with promising results, and have now acquired managerial commitment to develop a fully operational software environment for the prediction of product trends and component consumption. Because planners and analysts are already familiar with belief nets, they have an immediate hook with which to grasp the essential concepts underlying these fairly sophisticated forecasting techniques. In keeping with the fundamental approach of SCRAM, they believe the forecasts because they have a qualitative understanding of the basic terms and tools used to build them.

An important problem for Sun is to predict the consumption of its peripheral devices sold as options. Options are components which do not ship with systems such as Sparcs or Ultra machines. Customers typically buy memory and storage options when upgrading systems that they already own. Options are therefore difficult to predict because their consumption is not constrained by system shipments. Figure 5 shows a simple belief net for total memory consumption, a combination of non-option memory configured with current system shipments and pure option memory sold directly to the installed base of customers.

Algorithms such as those described in [Dagum and Galper et al, 1995] and [Dagum and Galper, 1995] can be used to learn the static model for total memory from historical data. Simultaneously, econometric methods (e.g., an additive model with fixed look-back) can be used to generate a dynamic statistical forecast for total memory. Given the inference for total memory at each time step, the statistical forecast at each time step, and some reasonable way of combining both probability distributions (cf. [Genest and Zidek, 1986]), the forecast for consumption of memory options becomes a matter of inference using any one of several available algorithms for exact or approximate inference. [Dagum and Galper, 1995] and [Dagum and Galper et al, 1995] provide a much fuller exposition of the technique than we can give here.

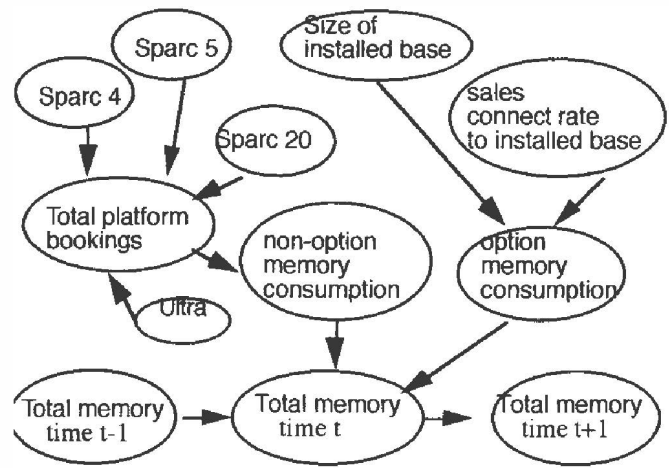

FIGURE 7. Belief net for total memory consumption.

As our pilot studies in this area convincingly demonstrate, the synthesis of belief net models learned from historical data with statistical forecasting methodologies gives planners and analysts a powerful tool for tracking the pace and pulse of the market. In particular, it allows them to test hypotheses by formulating a model such as the one displayed in Figure 7, fitting it to available data, and then seeing how the model performs against historical data. Approaches such as those developed in [Heckerman et al, 1995] are more sophisticated in that they seek to learn model structures—not just model parameters—from data. While this is clearly an attractive feature, one of the central premises of SCRAM is to get planners and analysts to articulate their beliefs using a simple, evocative representation. The search for credible models using the belief net inference/statistical forecasting approach becomes an experimental exercise where planners and analysts work together to balance ideas and intuitions against numerical signals in the data. It also provides a rigorous and surprisingly accurate means of forecasting phenomena such as memory options, products with demand and consumption patterns which are otherwise extremely difficult to predict.

## 5.0 Conclusions

At an invited talk at the Decision Analysis Colloquium at Stanford University, Howard Raiffa, one of the founders and luminaries of decision analysis (DA), stood before an audience of researchers, consultants, and students and proclaimed his dismay at DA's failure to exert any lasting, recognizable impact on current business practice. The critique was notable not just for its source — a man who helped lay the very foundations of the field — but also for its bluntness. When asked why DA had so few adherents, Raiffa offered one central explanation: too often, he said, DA had been presented as a collection of mathematical axioms and techniques instead of as an illuminating lens through which to view the world. He held up one of his recently published books, a thick tome filled with equations, theorems and proofs, and said, "See, this is exactly what I'm talking about!"

DA/UAI methods are appealing because they are based on crisp theories (e.g., subjective probability, utility) and because they maintain a strong attention to mathematical rigor. In my opinion they are most attractively distinguished, however, by their user interface. They provide practitioners and researchers with a rich set of tools, mostly graphical in nature, for expressing risk, complexity, and uncertainty — without necessarily requiring a huge amount of mathematical abstraction. To make themselves useful, other approaches seem to make sacrifices in rigor that are altogether too steep. Fuzzy logic, for example, starts from the premise that many of the concepts manipulated in the service of rational action are inherently fuzzy. While this may be true for *some* concepts, the fuzzy approach does not distinguish between those concepts and the less troublesome cases where the concepts themselves are relatively clear — the difficulty lies only in finding a compact way of expressing them and then tying them together to draw an inference or to make a decision.

My experience in the application of SCRAM for ODM indicates that, while better decision-making is of course one of its important benefits, another important benefit is that it forces decision participants to impose some structure on what only *appears* to be intrinsically fuzzy. The success of SCRAM at Sun demonstrates that graphical models such as belief nets and influence diagrams can be used on a fast, flexible basis to diagnose and depict the sources of disagreement in high-stakes business decisions. SCRAM analyses still retain all the mathematical rigor and complexity we need to solve our problems convincingly. In contrast to OR approaches, the results of models constructed according to SCRAM are more credible because decision participants build them using familiar terms, concepts, and pictures. Finally, the process first/tools second approach at Sun has led to a deeper understanding of the context and required functionalities of the tools for decision software support we are building.